\documentclass[11pt]{article}
\usepackage{amsfonts}
\usepackage{amsmath}
\usepackage{authblk}
\usepackage{booktabs}
\usepackage[margin=1in]{geometry}
\usepackage{graphicx}
\usepackage{makecell}
\usepackage{multirow}
\usepackage{physics}
\usepackage{subcaption}
\usepackage{threeparttable}
\usepackage[colorlinks=true,citecolor=blue]{hyperref}
\graphicspath{{figs/}}

\title{Fourier-MIONet: Fourier-enhanced multiple-input neural operators for multiphase modeling of geological carbon sequestration}
\author[1]{Zhongyi Jiang}
\author[1]{Min Zhu}
\author[1,2,*]{Lu Lu}
\affil[1]{Department of Statistics and Data Science, Yale University, New Haven, CT 06511, USA}
\affil[2]{Wu Tsai Institute, Yale University, New Haven, CT 06510, USA}
\affil[*]{Corresponding author. Email: lu.lu@yale.edu}
\date{}

\begin{document}

\maketitle

\begin{abstract}
Geologic carbon sequestration (GCS) is a safety-critical technology that aims to reduce the amount of carbon dioxide in the atmosphere, which also places high demands on reliability. Multiphase flow in porous media is essential to understand CO\textsubscript{2} migration and pressure fields in the subsurface associated with GCS. However, numerical simulation for such problems in 4D is computationally challenging and expensive, due to the multiphysics and multiscale nature of the highly nonlinear governing partial differential equations (PDEs). It prevents us from considering multiple subsurface scenarios and conducting real-time optimization. Here, we develop a Fourier-enhanced multiple-input neural operator (Fourier-MIONet) to learn the solution operator of the problem of multiphase flow in porous media. Fourier-MIONet utilizes the recently developed framework of the multiple-input deep neural operators (MIONet) and incorporates the Fourier neural operator (FNO) in the network architecture. Once Fourier-MIONet is trained, it can predict the evolution of saturation and pressure of the multiphase flow under various reservoir conditions, such as permeability and porosity heterogeneity, anisotropy, injection configurations, and multiphase flow properties. Compared to the enhanced FNO (U-FNO), the proposed Fourier-MIONet has 90\% fewer unknown parameters, and it can be trained in significantly less time (about 3.5 times faster) with much lower CPU memory ($<$ 15\%) and GPU memory ($<$ 35\%) requirements, to achieve similar prediction accuracy. In addition to the lower computational cost, Fourier-MIONet can be trained with only 6 snapshots of time to predict the PDE solutions for 30 years. Furthermore, we observed that Fourier-MIONet can maintain good accuracy when predicting out-of-distribution (OOD) data. The excellent generalizability of Fourier-MIONet is enabled by its adherence to the physical principle that the solution to a PDE is continuous over time. Moreover, the developed Fourier-MIONet makes it possible to solve the long-time evolution of geological carbon sequestration in a large-scale three-dimensional space accurately and efficiently.
\end{abstract}


\section{Introduction}

Multiphase flow in porous media is crucial in various industry and natural processes, such as geologic carbon sequestration (GCS)~\cite{pachauri2014climate}, enhanced oil recovery~\cite{aziz1979petroleum}, hydrogen production~\cite{aziz1979petroleum,hashemi2021pore}, and nuclear waste geological repositories~\cite{Amaziane2012}, which have high requirements for ensuring utmost reliability and safety. Numerical simulation is primarily used to solve the material and energy balances of the multiphase flow system~\cite{prosperetti2009computational,balachandar2010turbulent}, which involves highly nonlinear partial differential equations~\cite{orr2007theory} and requires high-resolution grids~\cite{doughty2010,wen2019co2} and multiscale and multiphysics modeling~\cite{khebzegga2020,meng2016localized}. Large-scale geological carbon sequestration projects involve optimization and decision-making tasks under subsurface uncertainties~\cite{Strandli2014,kitanidis_2015,kroker2022arbitrary,rehme2021b}, where a large number of forward simulations need to be conducted under various subsurface scenarios and realizations. For this purpose, conventional reservoir simulators are inefficient due to their high computational cost, whereas (physics-informed and/or data-driven) surrogate modeling has the potential to enable real-time predictions once trained on a reasonable amount of data from full-physics simulations~\cite{Tahmasebi2020,cardoso2009,razavi2012review,baza2015}.

Surrogate modeling based on deep neural networks that can learn solutions of PDEs is a promising alternative to traditional numerical simulation~\cite{ZHU2018,mo2019,TANG2020,wen2021ccs,liu2021uncertainty,wu2023adaptive}. Deep learning techniques have recently been used to solve subsurface flow and transport problems in carbon storage~\cite{wen2021ccs,wen2022u,wen2022acc}. The two most common deep learning techniques are data-driven learning~\cite{ZHU2018,mo2019,zhong2019,TANG2020,Wen2021Towards,ZHU2023116300,jiao2021one,liu2022predicting,fan2024accident,kandel2024data} and physics-informed learning~\cite{RAISSI2019,ZHU2019,lu2021deepxde,cai2021physics,fan2023deep}. The former has achieved accurate predictions even for high-dimensional multiphase flow problems~\cite{JIANG2021,TANG2021deep,wu2021physics,wen2021ccs}. However, data-driven learning requires a large amount of data for training. Physics-informed neural networks (PINNs)~\cite{karniadakis2021physics} and their extensions~\cite{pang2019fpinns,YU2022,WU2023} have shown promising applications in computational science and engineering~\cite{chen2020physics,lu2021physics,yazdani2020systems,daneker2022systems,song2024simulation,das2024reliability}. The physics-informed learning is usually implemented using artificial neural networks (ANNs)~\cite{kamrava2021, WANG2021}, thus requiring separate training if any new parameters or coefficients change~\cite{RAISSI2019}. Similar to traditional numerical simulations, physics-informed learning for solving geologic carbon storage problems can be computationally inefficient. In some cases, additional physical constraints or observations are needed to guarantee convergence~\cite{fuks2020,almajid2022prediction,fraces2021physics}.

In recent years, deep neural operators~\cite{lu2021learning,li2020fourier,Lu2022deeponetvsfno,zhu2023reliable,yin2022interfacing,goswami2022physics} have shown promising potential for many applications in computational science and engineering. Unlike traditional deep learning methods, deep neural operators are designed to learn operators between infinite-dimensional function spaces. Several numerical experiments have demonstrated the promise of deep neural operators, such as Fourier neural operator (FNO)~\cite{wen2022u}, for real-time prediction of reservoir simulation. Very recently, an extension of FNO, U-FNO, has achieved the state-of-the-art performance in highly heterogeneous geological formations for carbon sequestration problems~\cite{wen2022u}. However, U-FNO requires a couple of days for training in one GPU with about 16 GiB GPU memory and more than 100 GiB CPU memory, and it also needs a large training dataset and performs poorly on out-of-distribution (OOD) data.

In this work, we overcome these challenges of high computational cost and data efficiency. We propose a novel deep neural operator, Fourier-MIONet, based on the multiple-input deep neural operators (MIONet)~\cite{jin2022mionet}. MIONet is an extension of the deep neural operator (DeepONet)~\cite{lu2021learning, zhu2023reliable} and it enables several input functions defined on different domains. We further incorporate U-FNO in MIONet. Fourier-MIONet requires much less GPU and CPU memory and significantly reduces the computational cost without compromising prediction accuracy. Moreover, to simulate geological carbon sequestration for 30 years, the required number of time snapshots can be reduced from 24 to only 6. Additionally, Fourier-MIONet achieves better generalization for OOD scenarios. This finding could have very important implications for reducing monitoring costs and safety-critical applications.

This study presents a significant contribution to the field of GCS through the development of the Fourier-MIONet. By integrating MIONet and U-FNO, Fourier-MIONet offers an efficient and accurate method for simulating multiphase flow in porous media. The model substantially reduces computational costs, using 85\% less CPU memory and 64\% less GPU memory, and accelerates training by 3.5 times compared to U-FNO. Additionally, its ability to generalize from limited data ensures accurate long-term predictions of CO\textsubscript{2} migration and pressure fields. This contribution benefits real-time optimization and decision-making in GCS, a safety-critical technology, improving the feasibility and effectiveness of large-scale carbon sequestration projects aimed at mitigating climate change.

The paper is organized as follows. In Section~\ref{problemsetup}, we introduce the problem setup, including the governing equations of multiphase flow and the dataset generated by numerical simulation. In Section~\ref{Method}, after providing a brief overview of MIONet and U-FNO, we propose Fourier-MIONet. In Section~\ref{results}, we demonstrate the accuracy, efficiency, and reliability of Fourier-MIONet on multiphase flow prediction for gas saturation and pressure. We conclude the paper in Section~\ref{conclusions} and give a brief discussion of the 4D problem.

\section{Problem setup}
\label{problemsetup}

In this study, we consider the problem of multiphase flow in porous media. Specifically, we consider the application of underground carbon storage (also called geological carbon sequestration), which involves injecting supercritical CO\textsubscript{2} into underground geological formations for permanent storage. Geological carbon sequestration is one of the most important technologies for mitigating CO\textsubscript{2} emissions~\cite{benson2008co2, zhang2014mechanisms}.

\subsection{Multiphase flow}

The problem we are considering is multiphase, multicomponent  (CO\textsubscript{2}, brine) flow in porous media. Each component $\alpha$ satisfies the mass conservation equation:
$$\frac{\partial M^{\alpha}}{\partial t}=-\div\left(\vb{F}^{\alpha}|_{adv} + \vb{F}^{\alpha}|_{dif}\right)+q^{\alpha},$$
where $\vb{F}^{\alpha}|_{adv}$ is the advective mass flux, $\vb{F}^{\alpha}|_{dif}$ is the diffusive mass flux, $q^{\alpha}$ is the volumetric flow rate of the injection source, and $M^{\alpha}$ is the accumulation of mass given by
$$M^\alpha=\phi\sum_{p}S_p\rho_{p}X_{p}^\alpha.$$
Here, $\phi$ is the porosity, $S_p$ is the saturation of phase $p$, $X_{p}^{\alpha}$ is the mass fraction of component $\alpha$ in phase $p$, and $\rho_p$ is the density of phase $p$.

For simplicity, we do not consider molecular diffusion and hydrodynamic dispersion, and thus $\vb{F}^{\alpha}|_{dif} = 0$. The advective mass flux of component $\alpha$ is
$$\vb{F}^{\alpha}|_{adv}=\sum_{p}X_{p}^{\alpha}\rho_{p}\vb{u}_{p}.$$
Here, $\vb{u}_p$ is the Darcy velocity of phase $p$ as follows:
$$\vb{u}_{p}=-k\left(\nabla P_{p} - \rho_{p}\vb{g}\right)k_{rp}/\mu_{p},$$
where $k$ is the absolute permeability tensor, $P_p$ is the fluid pressure of phase $p$, $\vb{g}$ is the gravitational acceleration, $k_{rp}$ is the relative permeability of phase $p$, and $\mu_{p}$ is the viscosity of phase $p$. The fluid pressure for wetting phase $P_w$ or non-wetting phase $P_n$ is
$$P_n = P_w + P_c,$$
where $P_c$ is the capillary pressure.

\subsection{Dataset}

In this study, we use the open-source dataset from Wen et al.~\cite{{wen2022u}}, generated for the purpose of testing surrogate modeling algorithms. To generate the dataset, we consider that CO\textsubscript{2} is injected at a constant rate into a radially symmetrical system. The thickness of the reservoir ranges from 12.5 to 200 m and the radius of the reservoir is 100,000 m. Closed boundary condition is applied at the top and bottom of the reservoir, meaning the flux normal to the boundary is zero. To mimic an open reservoir condition, a very large simulation domain was used with no flow boundary condition. The system is solved numerically with the finite difference method by the simulator ECLIPSE (e300) for 30 years. 

In the dataset, several PDE parameters (i.e., network inputs) are randomly sampled to generate different PDE solutions (i.e., network outputs): gas saturation (SG) and pressure buildup (dP). These inputs can be categorized into two types (Table~\ref{table1}): space-dependent inputs (field inputs) and scalar inputs. Field inputs include the horizontal permeability map ($k_x$), vertical permeability map ($k_y$), and porosity map ($\phi$). Permeability maps $k_x$ are produced by the Stanford Geostatistical Modeling Software (SGeMS)~\cite{remy2009applied} based on different parameters, including correlation lengths in the vertical and radial directions, medium appearances, permeability mean and standard deviation. The permeability value ranges widely from 0.001 $\mathrm{mD}$ to 10000 $\mathrm{mD}$. Values of $k_x$ are binned into $n_{\text{aniso}}$ tropic materials, where each bin is assigned a randomly sampled anisotropy ratio ($k_x / k_y$). Then the vertical permeability map $k_y$ is computed by multiplication between $k_x$ and the anisotropy map. Porosity ($\phi$) is calculated by loosely correlated permeability map with a random Gaussian noise $\mathcal{N}(0, 0.001)$. 

\begin{table}[htbp]
    \centering
    \caption{\textbf{Summary of the branch network inputs.} The size of the field inputs is (96, 200). $\mathcal{V}[a,b]$ means that the input value ranges from $a$ to $b$.}
    \label{table1}
    \begin{tabular}{lllll}
        \toprule & Parameter & Notation & Distribution & Unit \\
        \midrule Field & \makecell[l]{Horizontal permeability field \\Material anisotropy ratio \\Porosity} & \makecell[l]{$k_x$\\$k_x / k_y$\\$\phi$} & \makecell[l]{$\mathcal{V}[0.001,10000]$\\$\mathcal{V}[1,150]$\\--} & \makecell[l]{mD\\--\\--} \\
        \midrule Scalar & \makecell[l]{Injection rate\\Initial pressure\\Iso-thermal reservoir temperature\\Irreducible water saturation\\Van Genuchten scaling factor\\Perforation top location\\Perforation bottom location} & \makecell[l]{$Q$\\$P_{\text {init }}$\\$T$\\$S_{wi}$\\$\lambda$\\$perf_{top}$\\$perf_{bottom}$} & \makecell[l]{$\mathcal{V}[0.2,2]$\\$\mathcal{V}[100,300]$\\$\mathcal{V}[35,170]$\\$\mathcal{V}[0.1,0.3]$\\$\mathcal{V}[0.3,0.7]$\\$\mathcal{V}[0, 200]$\\$\mathcal{V}[0, 200]$} & \makecell[l]{MT/y\\bar\\${ }^{\circ}\mathrm{C}$\\--\\--\\$\mathrm{m}$\\$\mathrm{m}$} \\
        \bottomrule
    \end{tabular}
\end{table}

The scalar inputs include the initial reservoir pressure at the top of the reservoir ($P_{\text{init}}$), reservoir temperature ($T$), injection rate ($Q$), capillary pressure scaling factor ($\lambda$), irreducible water saturation ($S_{wi}$), and the top and bottom locations of the perforation ($perf_{top}$ and $perf_{bottom}$). We note that the reservoir thickness $b$ is a variable randomly sampled in each case and determines the size of the field inputs, ranging from 12.5 to 200 m. When $b$ is thinner than 200 m, zero padding is used for the domain outside of the actual reservoir such that all inputs have the same shape. Then, $perf_{top}$ and $perf_{bottom}$ are randomly sampled within the range from 0 to $b$ of each case.

\begin{figure}[htbp]
    \centering
    \includegraphics[width=\textwidth]{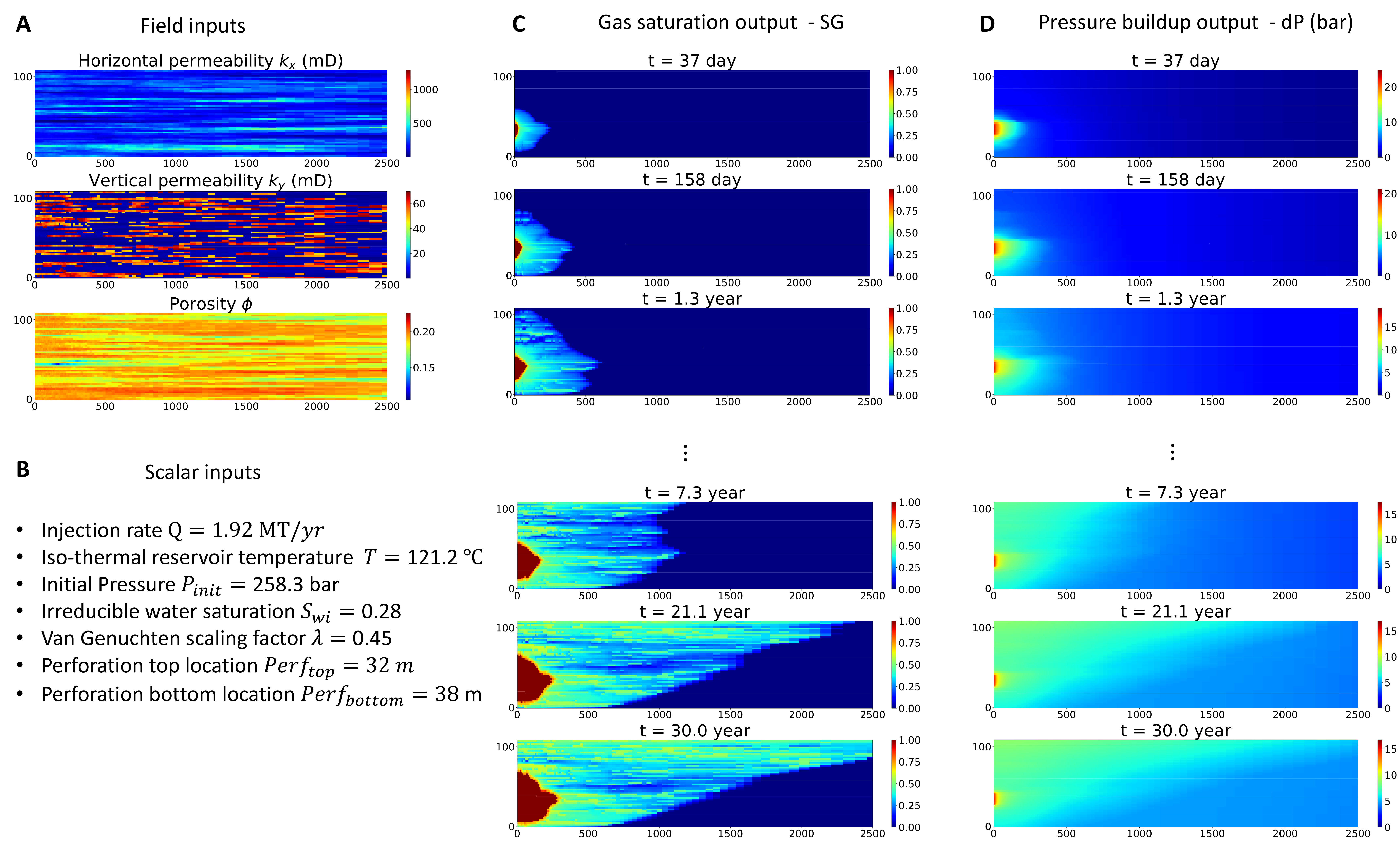}
    \caption{\textbf{An example of inputs and outputs.} (\textbf{A}) Example of field inputs. (\textbf{B}) Example of scalar inputs. (\textbf{C}) Gas saturation evolution at 6 out of 24 time snapshots. (\textbf{D}) Pressure buildup evolution at 6 out of 24 time snapshots.} 
    \label{Fig:case example}
\end{figure}

All the input and output fields in the dataset are saved in a resolution of (96, 200). The outputs are saved at 24 time snapshots from 1 day to 30 years: \{1 day, 2 days, 4 days, $\dots$ ,14.8 years, 21.1 years, 30 years\}. The training dataset includes 4500 cases, and the test dataset includes 500 cases. We show an example of inputs and outputs in Fig.~\ref{Fig:case example}.

\section{Methods}
\label{Method}

U-FNO has been developed to build a surrogate model for such dataset, and has shown great prediction accuracy. However, there are still some challenges remaining: the computational cost is high, as we have discussed in the introduction. Moreover, these existing methods always use different channels to predict the solutions at different times, which do not guarantee the continuity of solution over time. The recently developed framework of the multiple-input deep neural operators (MIONet) overcomes the challenges of computational cost and data efficiency, but MIONet by itself cannot easily provide highly accurate predictions. In this section, we first briefly introduce MIONet and U-FNO. Next, we propose Fourier-enhanced multiple-input neural operator (Fourier-MIONet) by combining the advantages of MIONet and U-FNO together for learning physical systems more effectively.

\subsection{MIONet}
\label{subsec:mionet}
MIONet was proposed by Jin et al.~\cite{jin2022mionet} for learning nonlinear operators mapping between function spaces. Based on the universal approximation theorem of Chen \& Chen~\cite{chen1995universal}, the vanilla DeepONet is defined for input functions on a single Banach space. MIONet extends the capability of DeepONet~\cite{lu2021learning} both theoretically and numerically from a single Banach space to multiple Banach spaces. MIONet can be understood as a multivariable regression method in the statistical regression area, but MIONet is different from other special regression models~\cite{chen2023modified, ma2015varying} which were designed to exploit the relationships among multiple variables.

We denote $n$ input functions by $v_i$ for $i=\{1, \dots, n\}$ with each defined on the domain $D_i \subset \mathbb{R}^{d_{i}}$:
$$
v_i: D_i \to \mathbb{R},
$$
and the output function by $u$ defined on the domain $D' \in \mathbb{R}^{d^{\prime}}$:
$$
u: D^{\prime} \ni \xi \mapsto u(\xi) \in \mathbb{R} .
$$
Then, the operator mapping from the input functions to the output function is 
$$
\mathcal{G}: (v_1, \dots, v_n)\mapsto u.
$$

\begin{figure}[htbp]
    \centering
    \includegraphics[scale=0.45]{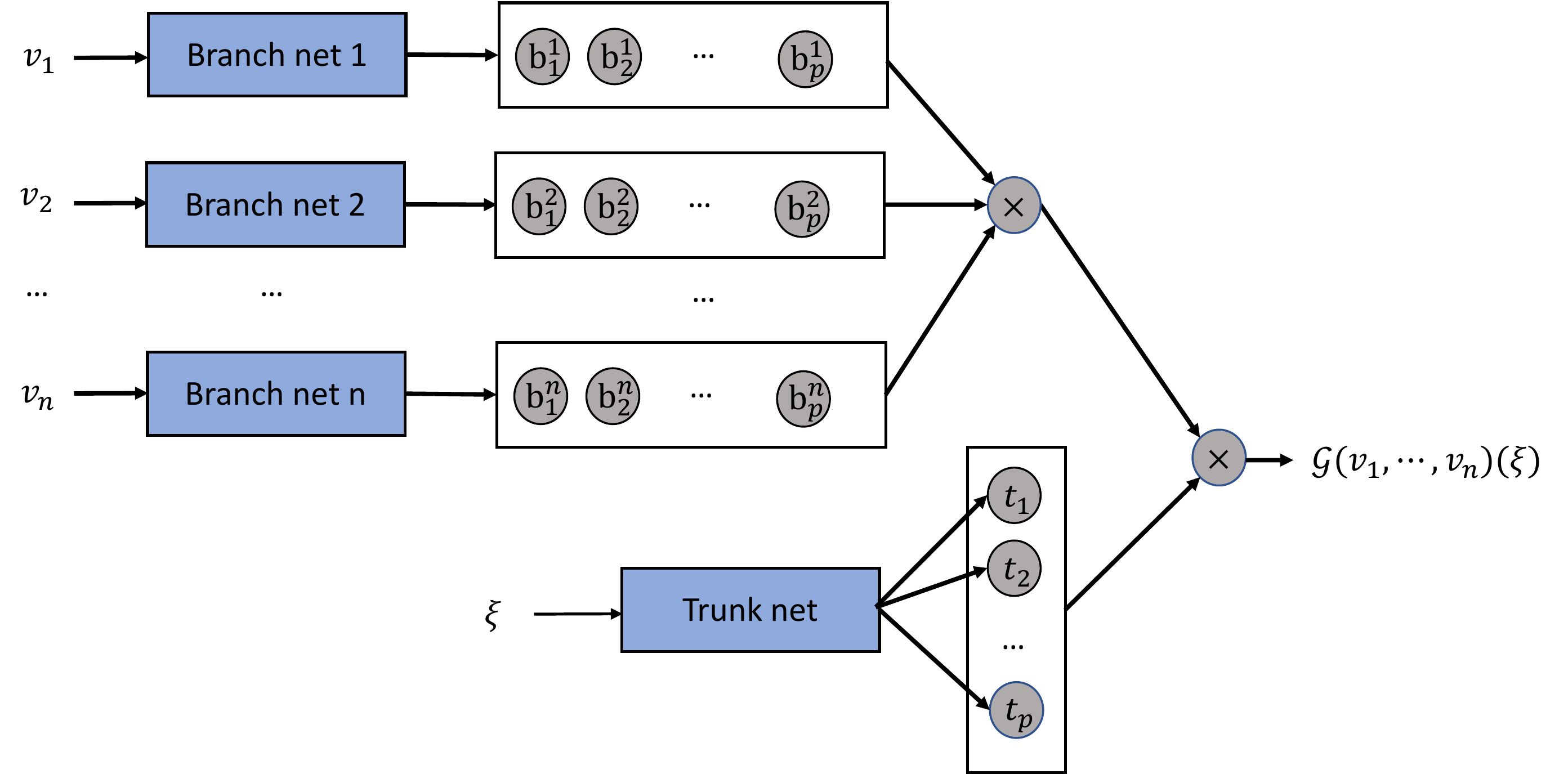}
    \caption{\textbf{Architecture of MIONet.} All the branch nets and the trunk net have the same
    number of outputs, which are merged together via the element-wise product.} \label{Fig1:MIONet}
\end{figure}

MIONet is developed  to learn the operator $\mathcal{G}$ by using $n$ independent branch nets and one trunk net (Fig.~\ref{Fig1:MIONet}). The $i$th branch net encodes the input function $v_i$, and the trunk net encodes the coordinates input $\xi$. The output of MIONet is computed as
$$
\mathcal{G}(v_1, v_2,\dots,v_n)(\xi)=\sum_{j=1}^p\underbrace{b^1_j(v_1)}_{\text{branch}_1} \times \underbrace{b^2_j(v_2)}_{\text{branch}_2}\cdots \times \underbrace{b^n_j(v_n)}_{\text{branch}_n} \times \underbrace{t_j(\xi)}_{\text{trunk}}+b_0,
$$
where $b_0 \in \mathbb{R}$ is a bias, $\left\{b_1^i, b_2^i, \ldots, b_p^i\right\}$ are the $p$ outputs of the branch net $i$, and $\left\{t_1, t_2, \ldots, t_p\right\}$ are the $p$ outputs of the trunk net. The choice of the branch nets and trunk net is problem dependent. For example, if the discretization of $v_i$ is on an equispaced 2D grid, then a convolution neural network can be used as the branch net $i$; if the discretization of $v_i$ is on an unstructured mesh, then a graph neural network can be used as the branch net $i$.

\subsection{U-FNO}

\begin{figure}[htbp]
    \centering
    \includegraphics[width=\textwidth]{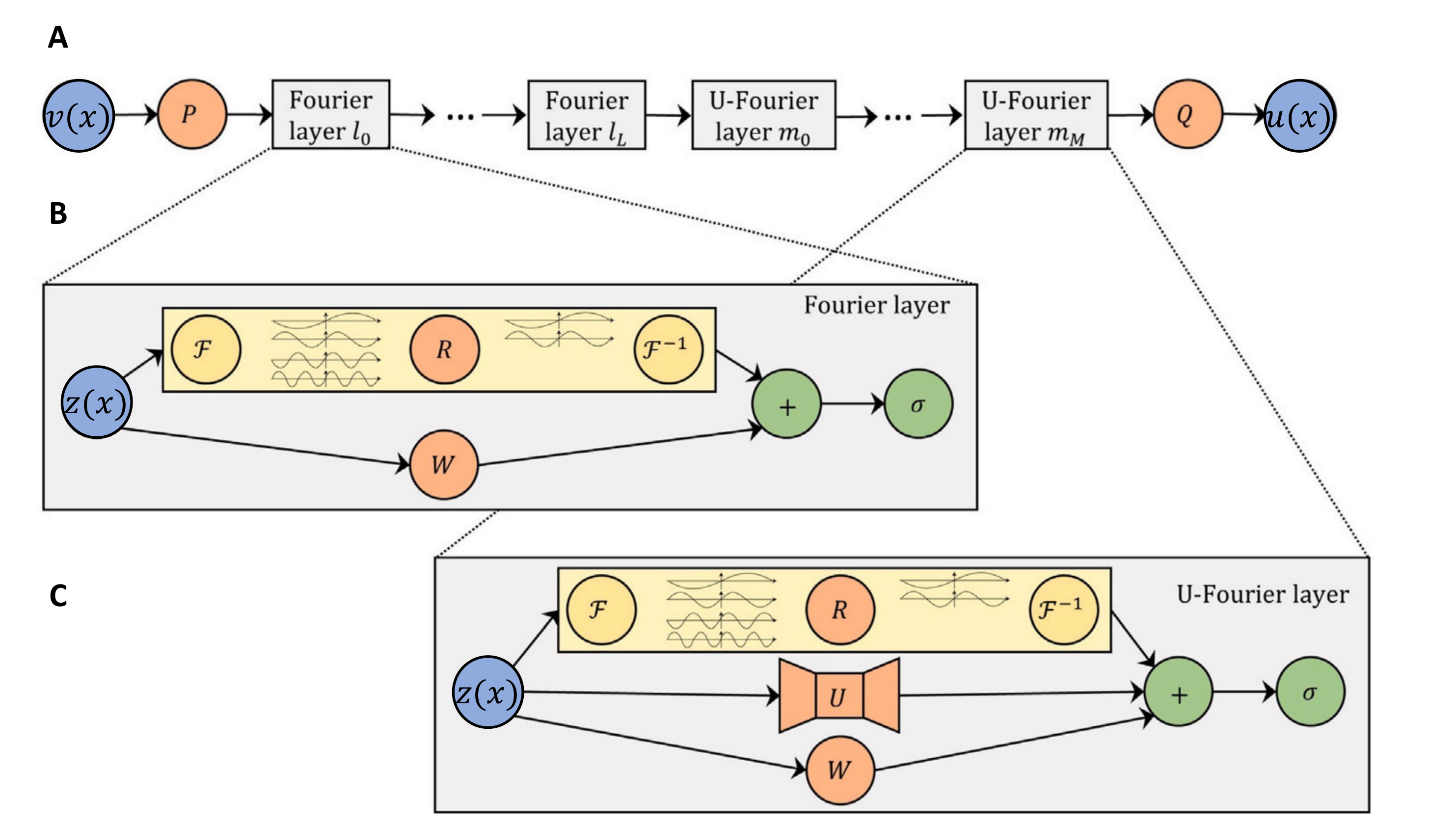}
    \caption{\textbf{Architecture of U-FNO.} (\textbf{A}) $v(x)$ is the input function, $P$ and $Q$ are fully connected neural networks, and $u(x)$ is the output function. (\textbf{B})  Inside each Fourier layer, $\mathcal{F}$ denotes the Fourier transform, $\mathcal{R}$ is a weight matrix, $\mathcal{F}^{-1}$ is the inverse Fourier transform, $W$ is an another weight matrix, and $\sigma$ is the activation function. (\textbf{C})  Inside each U-Fourier layer, $U$ denotes a U-Net block. Figure is adapted from~\cite{wen2022u}.} \label{Fig3:UFNO}
\end{figure}

U-FNO~\cite{wen2022u} is an extension of Fourier neural operator (FNO)~\cite{li2020fourier}, which computes convolutions in the Fourier space rather than physical space. U-FNO uses additional U-Net blocks~\cite{ronneberger2015u} in FNO. Different from DeepONet, U-FNO requires the input function $v(x)$ and output function $u(\xi)$ defined on the same equispaced mesh grid of the same domain. A schematic diagram of U-FNO is shown in Fig.~\ref{Fig3:UFNO}.

First, the input function $v(x)$ is lifted to a higher dimensional representation $z_{l_{0}}(x)$ by
$$
z_{l_{0}}(x)=P(v(x)),
$$
where $P$ is a linear transformation, which is often parameterized by a linear layer or a shallow  fully-connected neural network. Then, $L$ Fourier layers and $M$ U-Fourier layers are applied iteratively to $z_{l_0}$ and the final output is $z_{m_M}$. In the end, a local linear transformation $Q$ is applied by employing a shallow neural network to convert the $z_{m_M}$ to the output by
$$
u(x)=Q\left(z_{m_M}(x)\right).
$$

Each Fourier layer is defined by using the Fast Fourier Transform (FFT).
For the output of the $j$th Fourier layer $z_{l_j}$, we compute the following transform by 3D FFT $\mathcal{F}$ and inverse 3D FFT $\mathcal{F}^{-1}$:
$$
\mathcal{F}^{-1}\left(\mathcal{R}_{l_j} \cdot \mathcal{F}\left(z_{l_j}\right)\right),
$$
where $\mathcal{R}$ is a weight matrix. Moreover, a residual connection with a weight matrix $W_j$ is used to compute the output of the $(j + 1)$th Fourier layer $z_{l_{j+1}}$ as
$$
z_{l_{j+1}}=\sigma\left(\mathcal{F}^{-1}\left(\mathcal{R}_{l_j} \cdot \mathcal{F}\left(z_{l_j}\right)\right)+W_j \cdot z_{l_j}+b_{l_j}\right),
$$
where $\sigma$ is a nonlinear activation and $b_{l_j}$ is a bias.

U-Fourier layer is defined based on the Fourier layer and 3D U-Net block:
$$
z_{m_{k+1}}=\sigma\left(\mathcal{F}^{-1}\left(\mathcal{R}_{m_k} \cdot \mathcal{F}\left(z_{m_k}\right)\right)+U_{m_k}\left(  z_{m_k}\right)+W_{m_k} \cdot z_{m_k}+\boldsymbol{b}_{m_k}\right)
$$
where  $U_{m_k}$ is an U-Net block.

\subsection{Fourier-MIONet}

\begin{figure}[htbp]
    \centering
    \includegraphics[width=\textwidth]{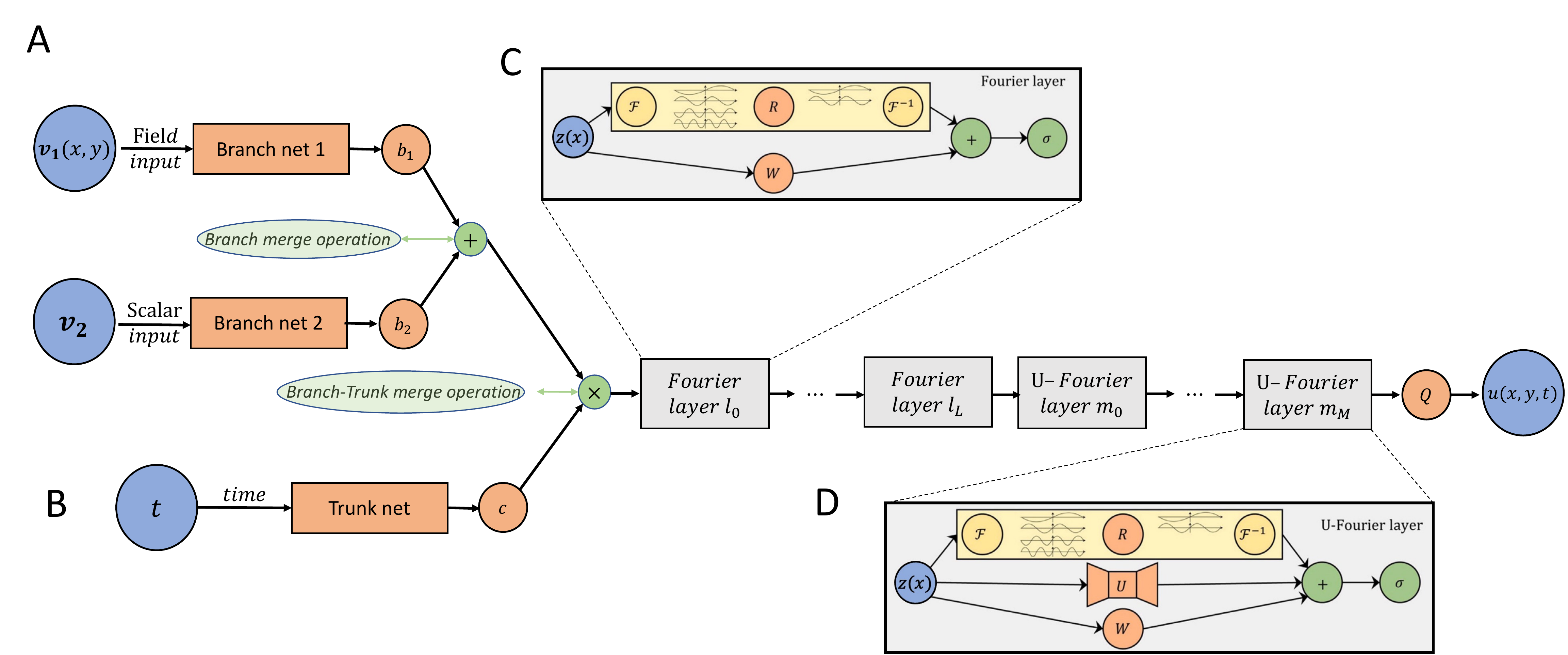}
    \caption{\textbf{Fourier-MIONet architecture.} (\textbf{A}) $\mathbf{v}_1$ is field input and $\mathbf{v}_2$ is scalar input. (\textbf{B}) The input $t$ is time. (\textbf{C}) Fourier layer. (\textbf{D}) U-Fourier layer.} \label{Fig4:FMIONet}
\end{figure}

Motivated by the discussion at the beginning of this section, here we develop a new version of MIONet, Fourier-MIONet, by leveraging MIONet and U-FNO. As we have discussed in Section~\ref{problemsetup}, we split the inputs into two parts: for all the field parameters denoted by $\mathbf{v_1}$ and for all the scalar parameters denoted by $\mathbf{v_2}$ (Fig.~\ref{Fig4:FMIONet}). The coordinates of the output function $u$ are $x$, $y$, and $t$, while $x$ and $y$ are space coordinates and $t$ is time. If we apply MIONet, the trunk net input includes $x$, $y$, and $t$, and the output is $u(x,y,t) \in \mathbb{R}$, while for U-FNO, the network output is $u(\cdot,\cdot,\cdot) \in \mathbb{R}^{96\times200\times24}$. In our developed Fourier-MIONet, we treat spatial and temporal coordinates separately. Specifically, the trunk input is only $t$, and the network output is $u(\cdot,\cdot,t) \in \mathbb{R}^{96\times200}$. 

In Fourier-MIONet, there are two branch nets to encode field inputs and scalar inputs, respectively (Fig.~\ref{Fig4:FMIONet}A). Let $\mathbf{b}_1$ and $\mathbf{b}_2$ be the outputs of the two branch nets, and $\mathbf{c}$ be the output of the trunk net. Two merger operations are used here: one to merge the two branch outputs (Branch merger operation) and one to merge the branch and trunk net outputs (Branch-Trunk merger operation). We choose the Branch merger operation as a point-wise summation:
$$
    \mathbf{b} = \mathbf{b}_1 + \mathbf{b}_2,
$$
where $\mathbf{b}$ is the output of the branch merger operation. We choose the Branch-Trunk merger operation as a point-wise multiplication used in vanilla MIONet:
$$
    z = \mathbf{b} \odot \mathbf{c},
$$
where $z$ is the output of the  merger operation. We note that ther are other choices of the merger operations, which will be studied in the future.

Next, we use U-FNO as a decoder to map from the hidden vector $z$ to the output $u$ by applying Fourier layers and U-Fourier layers iteratively (Figs.~\ref{Fig4:FMIONet}C and D). We note that we use 2D FFT and 2D U-Net in this U-FNO, rather than the 3D FFT and 3D U-Net used in the original U-FNO. In the end, a fully connected neural network $Q$ is applied to project the output of the last U-Fourier layer to the output.

\subsection{Training and evaluation}\label{training setup}
As mentioned in Section \ref{problemsetup}, our dataset  contains reservoirs with various thicknesses, and the domain outside of the reservoir is padded with zeros for both input and output. To accommodate for the variable reservoir thicknesses, we construct a mask for each data sample and only calculate the loss within the mask during training. In addition to the input variables described in Table \ref{table1}, we also supply the spatial grid information to the training as additional two field inputs.

Same as U-FNO~\cite{wen2022u}, we use $lp$-loss function:
$$
L(y, \hat{y})=\frac{\|y-\hat{y}\|_p}{\|y\|_p}+\beta \frac{\|\mathrm{d} y / \mathrm{d} r-\mathrm{d}\hat{ y} / \mathrm{d} r\|_p}{\|\mathrm{~d} y / \mathrm{d} r\|_p},
$$
where $y$ is the true solution and $\hat{y}$ is the prediction. $p$ is  the order of norm, and $\beta$ is a hyper-parameter. We choose $p=2$ and $\beta=0.5$. During training, the initial learning rate is specified to be 0.001, same as~\cite{wen2022u}, the learning rate gradually decreases with a constant rate.

U-FNO directly outputs gas saturation and pressure fields at all the 24 time steps during training and testing, i.e., the output size is (96, 200, 24). In each training step, we choose different cases and the number of cases (i.e., batch size) is denoted by $batch_{case}$. For Fourier-MIONet, in addition to $batch_{case}$, we can also choose different number of time snapshots of the trunk net input for training, and we denote this batch size as $batch_{time}$. Hence, when training Fourier-MIONet model, the output size is  (96, 200, $batch_{time}$).

To evaluate the performance of the trained network, we use $R^2$ and mean absolute error (MAE) as the metrics.

\section{Results}\label{results}

We demonstrate the accuracy, reliability, training efficiency, and data efficiency of the proposed Fourier-MIONet in this section. The network architecture for the Fourier-MIONet model is shown in Table~\ref{table3:FMIONet structure}. All experiments are performed on a workstation with an AMD Threadripper Pro 5955WX CPU and an NVIDIA GeForce RTX 3090 GPU. We implement the experiments by using the library DeepXDE~\cite{lu2021deepxde} and the code is available in GitHub at \url{https://github.com/lu-group/fourier-mionet-gcs}.

\begin{table}[htbp]
    \centering
    \caption{\textbf{Fourier-MIONet architecture}. In order to provide a fair comparison,  we adopt a similar structure for U-FNO as in~\cite{wen2022u}. The ``Padding'' denotes a padding operation. ``Linear'' denotes a linear transformation. ``Fourier2d'' denotes the 2D Fourier transform. ``Conv1d'' denotes 1D convolution. ``UNet2d'' denotes a 2D U-Net. $C$ represents $batch_{case}$. $T$ represents $batch_{time}$. In this model, the number of total parameters is 3,685,325.}
    \label{table3:FMIONet structure}
    \small
    \begin{tabular}{llll}
        \toprule &  & Operation & Output shape\\
        \midrule Branch net & \makecell[l]{Branch net 1\\Branch net 2} & \makecell[l]{Padding/Linear\\FNN} & \makecell[l]{($C$, 104, 208, 36)\\($C$, 36)}\\
        \midrule Branch merger operation &  & Point-wise summation & ($C$, 104, 208, 36)\\
        \midrule Trunk net &  & FNN & ($T$, 36)\\
        \midrule Branch-Trunk merger operation &  & Point-wise multiplication & ($C\times T$, 104, 208, 36)\\
        \midrule Merger net & \makecell[l]{Fourier 1\\Fourier 2\\Fourier 3\\U-Fourier 1\\U-Fourier 2\\U-Fourier 3\\Projection 1\\Projection 2\\De-padding\\Reshape} & \makecell[l]{Fourier2d/Conv1d/Add/ReLU\\Fourier2d/Conv1d/Add/ReLU\\Fourier2d/Conv1d/Add/ReLU\\Fourier2d/Conv1d/UNet2d/Add/ReLU\\Fourier2d/Conv1d/UNet2d/Add/ReLU\\Fourier2d/Conv1d/UNet2d/Add/ReLU\\Linear/ReLU\\Linear\\$-$\\$-$} & \makecell[l]{($C\times T$, 104, 208, 36)\\($C\times T$, 104, 208, 36)\\($C\times T$, 104, 208, 36)\\($C\times T$, 104, 208, 36)\\($C\times T$, 104, 208, 36)\\($C\times T$, 104, 208, 36)\\($C\times T$, 104, 208, 128)\\($C\times T$, 104, 208, 1)\\($C\times T$ , 96, 200, 1)\\($C$, $T$, 96, 200)}\\
        \bottomrule
    \end{tabular}
\end{table}

\subsection{Gas saturation} \label{sg}

We apply Fourier-MIONet to learn gas saturation. As we have discussed in Section~\ref{training setup}, in Fourier-MIONet, we have the flexibility to select different values of the $batch_{case}$ and $batch_{time}$, while U-FNO only allows for $batch_{case}$ (i.e., $batch_{time}$ can only be chosen as 24). Then, we first select $batch_{time}$ as 24 (i.e., full batch for the time input). The comparison between Fourier-MIONet and U-FNO is shown in the first two rows of Table~\ref{table2:detail main results}. First of all, the prediction accuracy between U-FNO and Fourier-MIONet is almost the same in terms of $R^2$ and MAE. Compared with U-FNO, Fourier-MIONet only has 1\% lower in the $R^2$. Compared to U-FNO with 33 million parameters, Fourier-MIONet only has 10\% trainable parameters (about 3 million). During training, U-FNO requires 103 GiB CPU memory and 15.9 GiB GPU memory, but Fourier-MIONet only needs 15 GiB CPU memory and 12.8 GiB GPU memory. Moreover, Fourier-MIONet is much faster to train, and the training time per epoch of Fourier-MIONet is 50\% less than U-FNO.

\begin{table}[htbp]
    \scriptsize
    \centering
    \caption{\textbf{Comparison between Fourier-MIONet and U-FNO.} We choose $batch_{case}$ as 4 for all the experiments. U-FNO uses all 24 time snapshots during training.}
    \label{table2:detail main results}
    \begin{tabular}{lccccccccc}
        \toprule & \makecell[c]{$batch_{time}$}  & \makecell[c]{No. of \\parameters} & \makecell[c]{CPU \\memory\\(GiB)} & \makecell[c]{GPU\\ memory\\(GiB)}& \makecell[c]{Training time\\per epoch\\  (second)} & \makecell[c]{Minimum \\epochs\\needed} & \makecell[c]{Training \\time\\ (hour)} & $R^2$ & MAE \\
        \midrule U-FNO & -- & 33,097,829 & 103 & 15.9  & 1535 & 100 & 42.6 & 0.992$\pm0.001$ & 0.0031$\pm0.0002$\\
        \midrule & 24 (Full) & 3,685,325 & 15 & 12.8  & 730 & 77$\pm13$ & 15.7$\pm2.7$ & 0.982$\pm0.002$ & 0.0050$\pm0.0002$ \\[1ex]
        & 20 & 3,685,325 & 15 & 11.0 & 742 & 66$\pm8$ & 13.7$\pm1.5$ & 0.979$\pm0.002$ & 0.0054$\pm0.0004$ \\[1ex]
        & 16 & 3,685,325 & 15 & 9.1 & 757 & 68$\pm13$ & 14.2$\pm2.8$ & 0.984$\pm0.001$ & 0.0046$\pm0.0002$ \\[1ex]
        & 12 & 3,685,325 & 15 & 7.3 & 808 & 56$\pm6$ & 12.6$\pm0.7$ & 0.984$\pm0.001$ & 0.0047$\pm0.0005$ \\[1ex]
        FMIONet & 8 & 3,685,325 & 15 & 5.6 & 928 & 48$\pm5$ & 12.3$\pm1.2$ & 0.985$\pm0.001$ & 0.0040$\pm0.0004$ \\[1ex]
        & 4 & 3,685,325 & 15 & 3.8 & 1339 & 73$\pm9$ & 27.2$\pm3.4$ & 0.987$\pm0.001$ & 0.0036$\pm0.0002$ \\[1ex]
        & 3 & 3,685,325 & 15 & 3.4 & 1609 & 54$\pm8$ & 24.1$\pm4.0$ & 0.987$\pm0.002$ & 0.0037$\pm0.0002$ \\[1ex]
        & 2 & 3,685,325 & 15 & 3.0 & 2195 & 44$\pm4$ & 27.0$\pm2.7$ & 0.988$\pm0.001$ & 0.0035$\pm0.0001$ \\[1ex]
        & 1 & 3,685,325 & 15 & 2.6 & 3832 & 39$\pm5$ & 41.8$\pm4.2$ & 0.987$\pm0.001$ & 0.0033$\pm0.0003$ \\
        \bottomrule
    \end{tabular}
\end{table}

\begin{figure}[htbp]
    \centering
    \includegraphics[width=\textwidth]{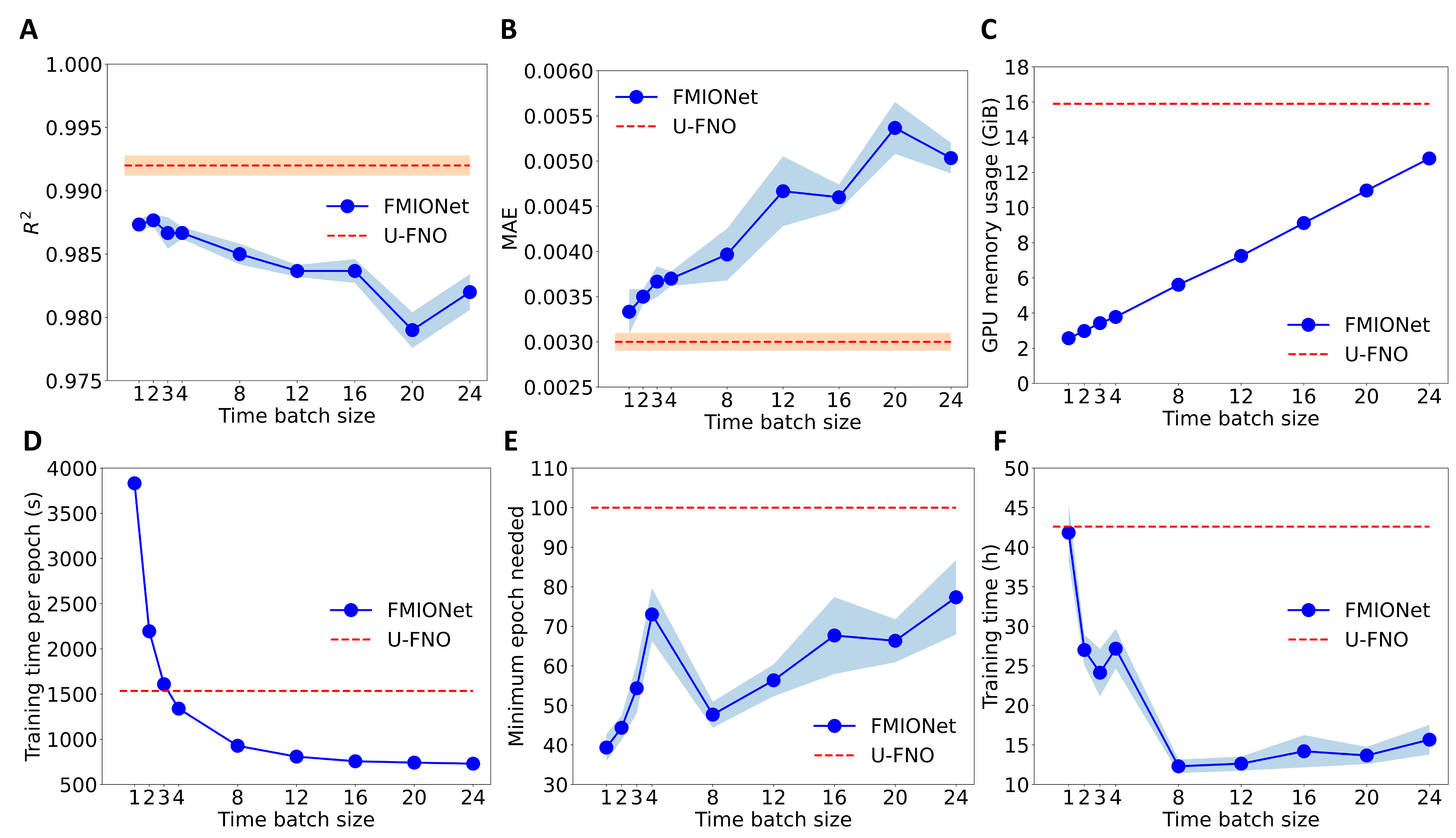}
    \caption{\textbf{Effect of $batch_{time}$ for Fourier-MIONet.} (\textbf{A}) $R^2$. (\textbf{B}) MAE. (\textbf{C}) GPU memory usage. (\textbf{D}) Training time per epoch. (\textbf{E}) Minimum number of training epochs needed. (\textbf{F}) Total training time. The solid blue curves and shaded regions represent the mean and one standard deviation of 3 runs of Fourier-MIONet. The precise values in this figure can be found in Table~\ref{table2:detail main results}. U-FNO does not have time batch size.}
    \label{6subplots}
\end{figure}

As Fourier-MIONet has the flexibility of using different values of $batch_{time}$, here we investigate the effect of $batch_{time}$ and determine the optimal choice of $batch_{time}$. The range of $batch_{time}$ is from 24 to 1. First, the $batch_{time}$ has little effect on the final accuracy (Figs.~\ref{6subplots}A and B), unless the $batch_{time}$ is very small (e.g., smaller than 4). In terms of computational cost, the GPU memory requirement decreases approximately linearly with respect to $batch_{time}$ (Fig.~\ref{6subplots}C). However, training time per epoch of the Fourier-MIONet increases when $batch_{time}$ decreases. In general, fewer epochs are needed for training neural networks when smaller batch size is used, which is also observed in Fourier-MIONet. Hence, for smaller batch size, less epochs are needed for the training to converge. In Fig.~\ref{6subplots}E, we show the minimum epoch needed for different $batch_{time}$. Since training time per epoch and minimum epoch needed have the opposite trend, the total training time has an approximate U-shaped curve (Fig.~\ref{6subplots}F). The minimum training time occurs at $batch_{time}$ equal to 8. Therefore, to achieve the fastest training, we use $batch_{time}$ as 8 in the following study.

\begin{figure}[htbp]
    \centering
    \includegraphics[width=\textwidth]{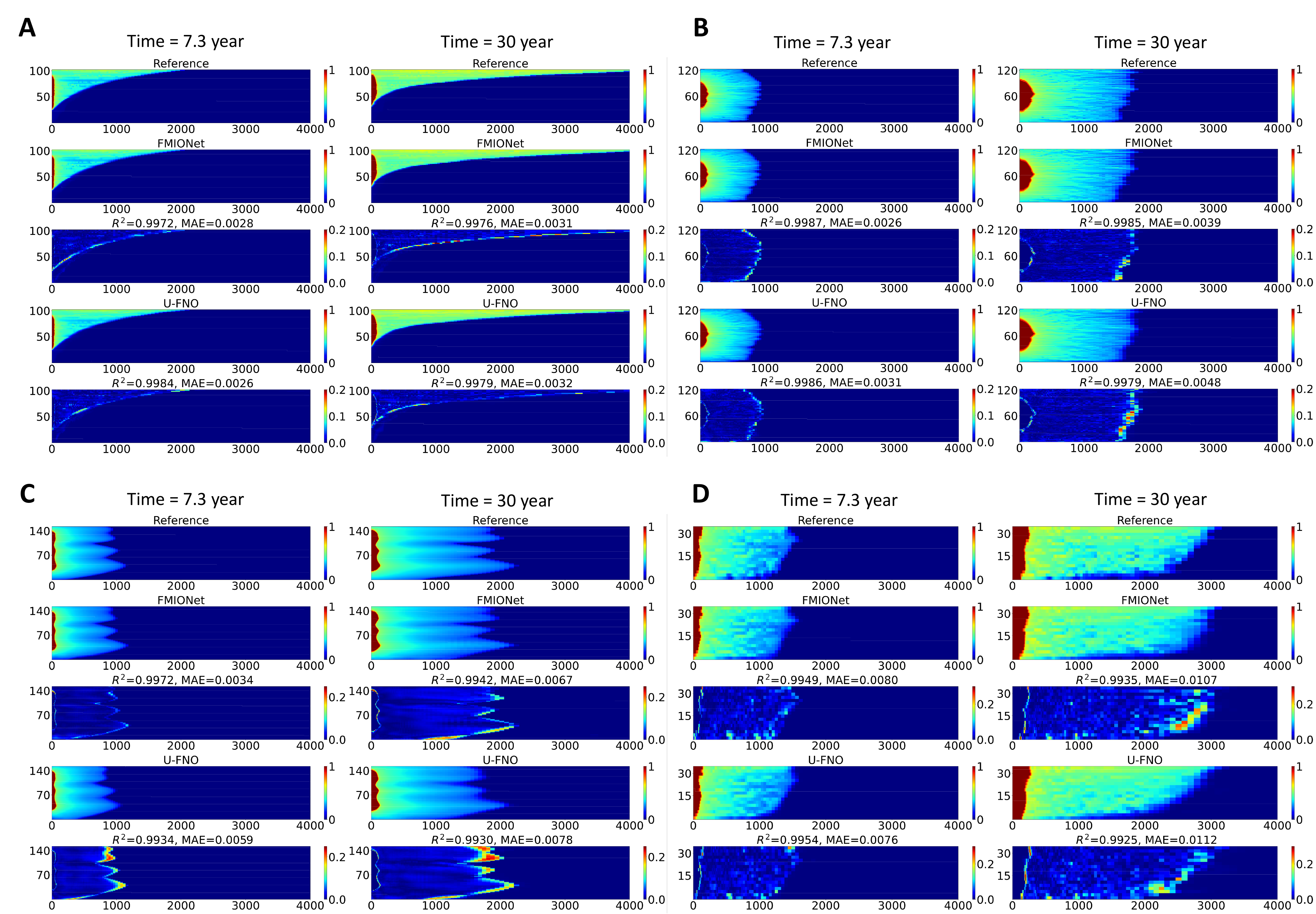}
    \caption{\textbf{Four testing examples of gas saturation.} For each example, we show the reference solution and the predictions and errors of Fourier-MIONet and U-FNO at 7.3 year and 30 year. The $R^2$ and MAE error are also computed for each prediction and shown in the figure.} 
    \label{4 sg cases}
\end{figure}

The accuracy between Fourier-MIONet (with $batch_{time}$ as 8) and U-FNO is almost the same and 4 testing examples of Fourier-MIONet are shown in Fig.~\ref{4 sg cases}. GPU memory of Fourier-MIONet is 5.69 GiB (Table~\ref{table2:detail main results}) which only accounts for 35\% of the U-FNO. Total training time of Fourier-MIONet is 12.3 hours (less than 30\% of U-FNO). Hence, Fourier-MIONet is much more computationally efficient.

In addition to the training cost, we also summarize the computational cost during testing (Table~\ref{table4:inference comparison}). Compared with U-FNO, Fourier-MIONet needs much less CPU memory (33.3\%) and GPU memory (50\%) during the testing. For each case, Fourier-MIONet requires 0.041 s, while U-FNO requires 0.075 s, so Fourier-MIONet is about 1.8 times faster.

\begin{table}[htbp]
    \centering
    \caption{\textbf{Computational cost of Fourier-MIONet and U-FNO during testing.} The inference time is computed by taking the average of 500 test cases.}
    \label{table4:inference comparison}
    \begin{tabular}{lcccc}
        \toprule & CPU memory (GiB) & GPU memory (GiB) & Inference time (s) \\
        \midrule U-FNO & 15.3 & 7.1 & 0.075\\
        FMIONet & 5.1 & 3.5 & 0.041\\ 
        \bottomrule
    \end{tabular}
\end{table}

\subsection{Inference for unseen time}

We also test the generalizability of Fourier-MIONet at unseen training time, i.e., only partial time snapshots are used for training. We recall that in the dataset for each case, there are 24 time snapshots. Here, we choose to test three cases with different amounts of training dataset: 50\%, 33\%, and 25\%.

\begin{figure}[htbp]
    \centering
    \includegraphics[width=.95\textwidth]{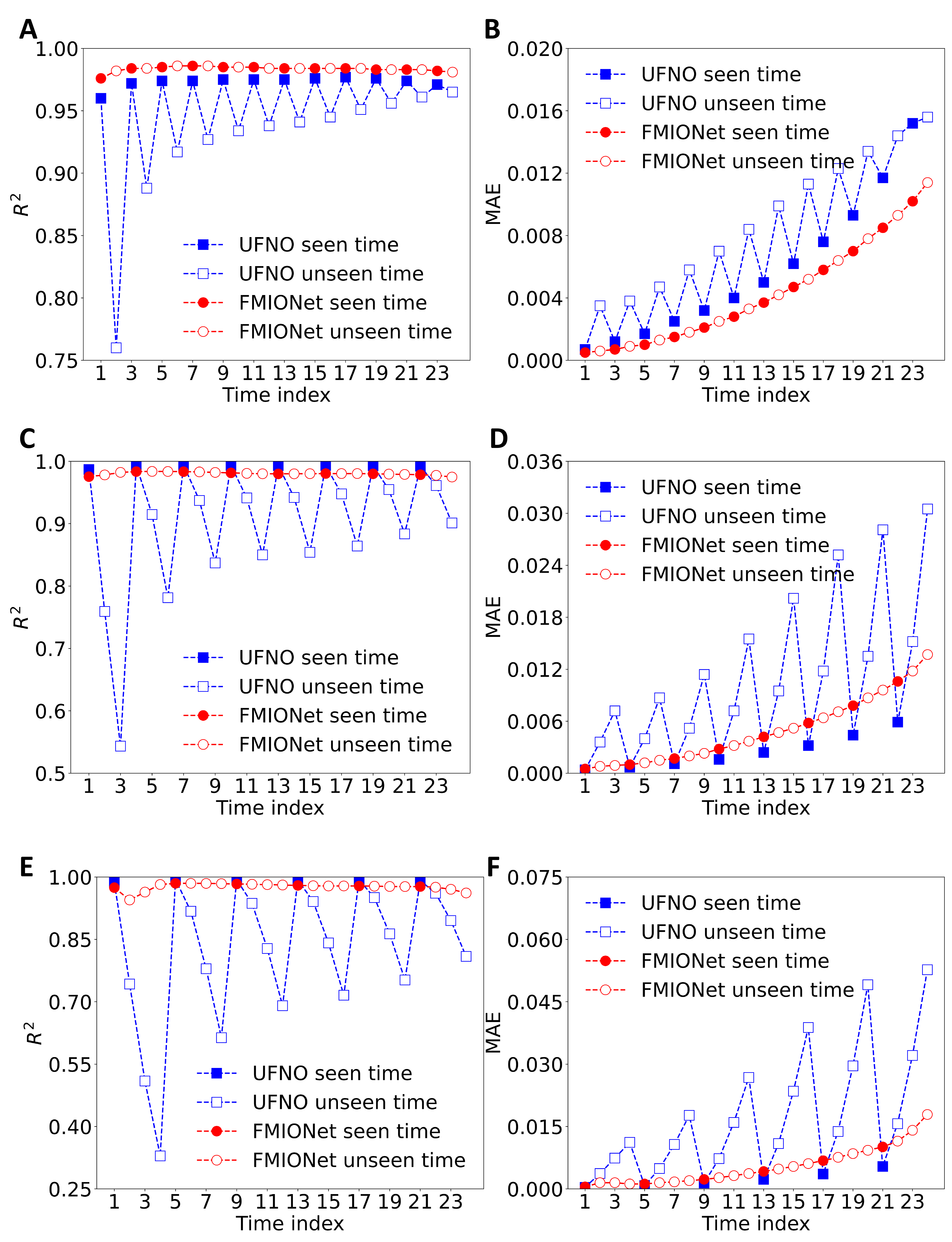}
    \caption{\textbf{Three different test cases for unseen time.} (\textbf{A} and \textbf{B}) Training at time index $\{1, 3, 5,7,9,11,13,15,17,19,21,23\}$. (\textbf{C} and \textbf{D}) Training at time index $\{1, 4, 7, 10, 13, 16, 19, 22\}$. (\textbf{E} and \textbf{F}) Training at time index $\{1, 5, 9, 13, 17, 21\}$.} 
    \label{Fig7:three test modes}
\end{figure}

In the first case, we use 50$\%$ of the time snapshots: $\{1, 3, 5,7,9,11,13,15,17,19,21,23\}$. As shown in Figs.~\ref{Fig7:three test modes}A and B, Fourier-MIONet has good performance at both seen time index and the unseen time index, and there is nearly no accuracy loss for the unseen time. However, U-FNO only performs well at seen time index, and a significant accuracy drop can be seen at the unseen time index.

The same phenomenon has also been observed for the other two cases of 33\% and 25\% training data (Figs.~\ref{Fig7:three test modes}C, D, E and F). We note that the behavior of U-FNO is typical for deep learning models, because it is usually difficult to generalize to unseen times by training with extremely sparse time snapshots. However, Fourier-MIONet shows good performance in both interpolation and extrapolation. For example in Figs.~\ref{Fig7:three test modes}E and F, the training time snapshots are 1 day, 11 day, 53 day, 226 day, 2.6 year, and 10.4 year, and the testing time snapshots include both interpolation (from 2 day to 7.3 year) and extrapolation (14.8 year, 21.1 year, and 30.0 year). The $R^2$ and MAE at 30.0 years are 0.9616 and 0.0179. Here, Fourier-MIONet can generalize well because we use time as the input of the trunk net, which guarantees that the prediction of gas saturation is continuous with respect to time, i.e., Fourier-MIONet obeys the physics law that gas saturation is continuous over, showcasing a crucial advance in ensuring the reliability of our model under varied conditions. In contrast, U-FNO uses different channels to predict gas saturation at different time, which does not guarantee the continuity of solution over time, weakening generalization ability for out-of-distribution (OOD) scenarios.

\subsection{Nonuniform sampling of training data}

To further improve the accuracy of using 25\% training data (i.e., six time snapshots) in the previous section, we propose to train the network using nonuniform time snapshots. The 24 time snapshots in the datasets corresponds to 1 day, 2 day, 4 day, 7 day, 11 day, 17 day, 25 day, 37 day, 53 day, 77 day, 111 day, 158 day, 226 day, 323 day, 1.3 year, 1.8 year, 2.6 year, 3.6 year, 5.2 year, 7.3 year, 10.4 year, 14.8 year, 21.1 year, and 30.0 year. Specifically, we design 6 different nonuniform sampling cases as listed in Table~\ref{6 sampling metrics}. The $R^2$ for each case can be found in Table~\ref{6 sampling metrics}.

\begin{table}[htbp]
    \centering
    \caption{\textbf{Six nonuniform sampling cases.}}
    \label{6 sampling metrics}
    \begin{tabular}{lccccc}
        \toprule
        &Case & Seen time snapshots & $R^2$\\ 
        \midrule &A &$\{$1d, 7d, 53d, 323d, 5.2y, 30y$\}$ & 0.9793$\pm$0.0247 \\[1ex]
        & B &$\{$1d, 4d, 53d, 323d, 5.2y, 30y$\}$ & 0.9809$\pm$0.0049\\[1ex]
        &C & $\{$1d, 4d, 37d, 226d, 5.2y, 30y$\}$ & 0.9821$\pm$0.0063\\[1ex]
        &D & $\{$1d, 4d, 37d, 226d, 3.6y, 30y$\}$ & 0.9807$\pm$0.0092\\[1ex]
        &E & $\{$1d, 4d, 37d, 323d, 5.2y, 30y$\}$ & 0.9793$\pm$0.0040\\[1ex]
        &F & $\{$1d, 4d, 37d, 323d, 7.3y, 30y$\}$ & 0.9773$\pm$0.0048\\
        \bottomrule
    \end{tabular}
\end{table}

\begin{figure}[htbp]
    \centering
    \includegraphics[width=\textwidth]{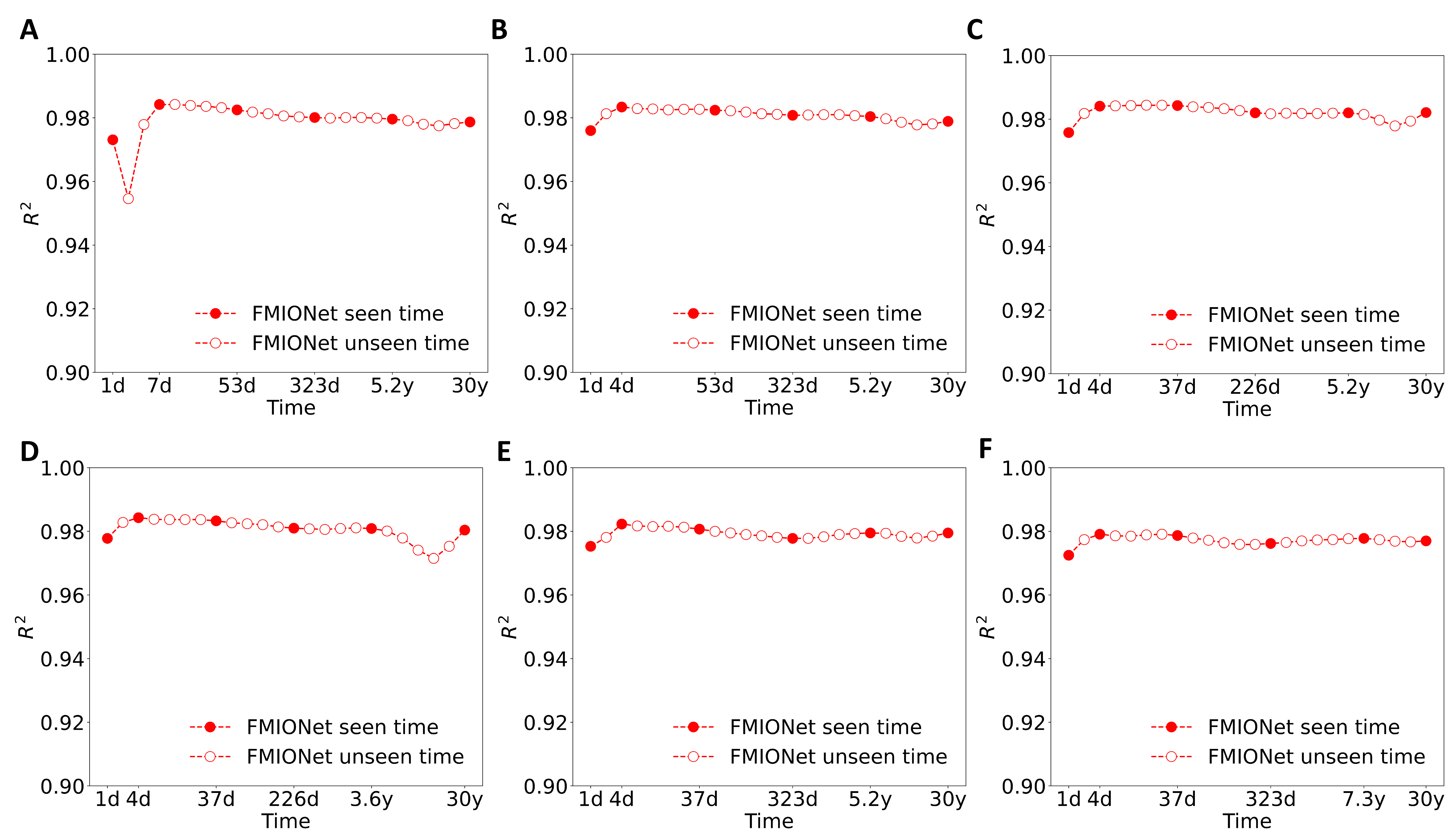}
    \caption{\textbf{Six nonuniform sampling of time.} (\textbf{A}) Training at time snapshots: $\{$1d, 7d, 53d, 323d, 5.2y, 30y$\}$. (\textbf{B}) Training at time snapshots: $\{$1d, 4d, 53d, 323d, 5.2y, 30y$\}$. (\textbf{C}) Training at time snapshots: $\{$1d, 4d, 37d, 226d, 5.2y, 30y$\}$. (\textbf{D}) Training at time snapshots: $\{$1d, 4d, 37d, 226d, 3.6y, 30y$\}$. (\textbf{E}) Training at time snapshots: $\{$1d, 4d, 37d, 323d, 5.2y, 30y$\}$. (\textbf{F}) Training at time snapshots: $\{$1d, 4d, 37d, 323d, 7.3y, 30y$\}$.}
    \label{6 sampling modes}
\end{figure}

For all these 6 cases, the $R^2$ of all time snapshots is  shown in Fig.~\ref{6 sampling modes}. Since the dynamics of gas saturation changes fast at the beginning, more data need to be chosen at the very first few time snapshots, otherwise, predictions will become worse. For example, if we choose the second time snapshots at 7 day (Fig.~\ref{6 sampling modes}A), the accuracy at the time 2 day will drop significantly. For later time, we need to choose time snapshots relatively uniformly. As shown in Fig.~\ref{6 sampling modes}D, there are 5 unseen time snapshots between the last two points, which makes the prediction of the unseen time with much lower accuracy. Hence, by choosing dense time snapshots at the beginning and relatively uniform snapshots between later times, Fourier-MIONet can achieve good prediction accuracy at both seen time and unseen time with only 6 time snapshots for training (Fig.~\ref{6 sampling modes}B, C, E and F).

\subsection{Pressure buildup}

\begin{figure}[htbp]
    \centering
    \includegraphics[width=\textwidth]{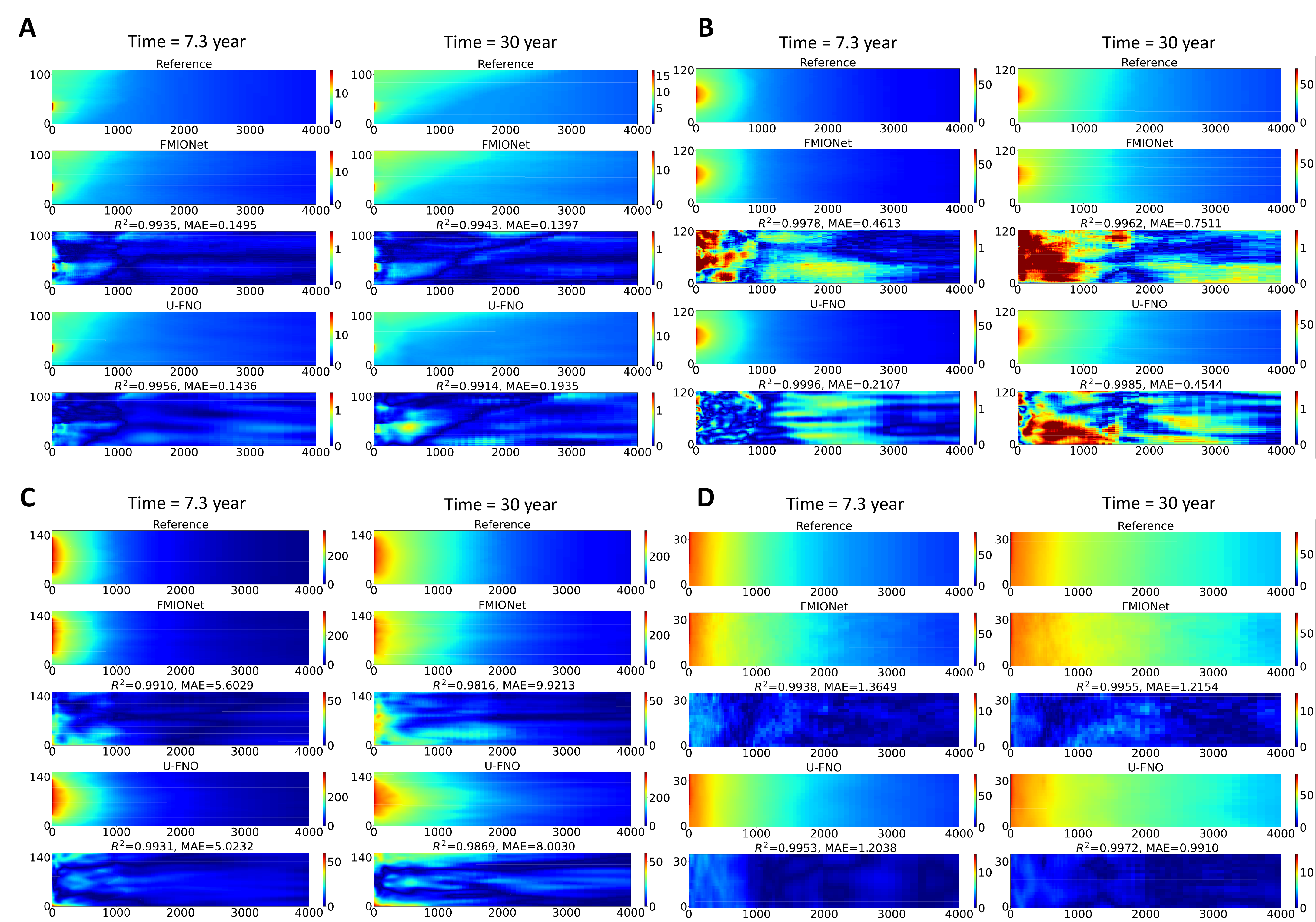}
    \caption{\textbf{Four testing examples of pressure buildup.} The reference solution and the predictions and errors of Fourier-MIONet and U-FNO at 7.3 year and 30 year. The $R^2$ and MAE error are also computed for each prediction and shown in the figure.} 
    \label{4 dp cases}
\end{figure}

When it comes to the prediction of pressure buildup, Fourier-MIONet also achieves good accuracy. We follow the same setup as U-FNO and use a smaller network for pressure buildup. Specifically, we only apply one Fourier layer and one U-Fourier layer in the merge net, and other parts of the network architecture are the same as the one in the gas saturation problem. Based on Section~\ref{sg}, we choose $batch_{time}$ as 8. The testing $R^2$ of Fourier-MIONet is 0.986. We show 4 examples of the prediction and absolute error with the corresponding $R^2$ and MAE in Fig.~\ref{4 dp cases}.

\subsection{Discussions about baseline network architectures}
\label{subsec:vanilla_mionet}
In this study, we have demonstrated good performance of Fourier-MIONet, and in this section, we show the necessity of using Fourier and U-Fourier layers as the merger net by studying two baseline networks. One baseline is the vanilla MIONet, which uses an inner product to construct the output (Section~\ref{subsec:mionet}). Another baseline is MIONet-FNN, where the dot product in the vanilla MIONet is replaced by a fully-connected neural network (FNN) as the merger net~\cite{mao2023ppdonet,seidman2022nomad}. The network architectures of the vanilla MIONet and MIONet-FNN are in Tables~\ref{mionet_structure} and \ref{mionet-FNN_structure}, respectively. 

The testing $R^2$ of vanilla MIONet for both gas saturation and pressure buildup is the worst among these three models (Table~\ref{mionet and mionet-fnn result}) due to the simple dot product in merger operation. By using FNN as the merger net, MIONet-FNN performs better than vanilla MIONet. Fourier-MIONet achieves the best accuracy by employing Fourier and U-Fourier layers as a decoder. Fourier layers with truncated modes can effectively capture low-frequency components in the data, while U-Nets are designed to capture both local and global features.

\begin{table}[htbp]
\centering
\caption{{\textbf{Accuracy of vanilla MIONet, MIONet-FNN and Fourier-MIONet.}}}
\label{mionet and mionet-fnn result}
{
\begin{tabular}{cc|c}
\toprule
 &  & $R^2$ \\   
\midrule
\multirow{2}{*}{Gas saturation} & Vanilla MIONet & 0.948 \\
& MIONet-FNN & 0.971 \\
& Fourier-MIONet & 0.985 \\
\midrule
\multirow{2}{*}{Pressure buildup} & Vanilla MIONet & 0.961 \\
& MIONet-FNN & 0.979 \\
& Fourier-MIONet & 0.986 \\
\bottomrule
\end{tabular}}
\end{table}

\begin{figure}[htbp]
    \centering
    \includegraphics[width=\textwidth]{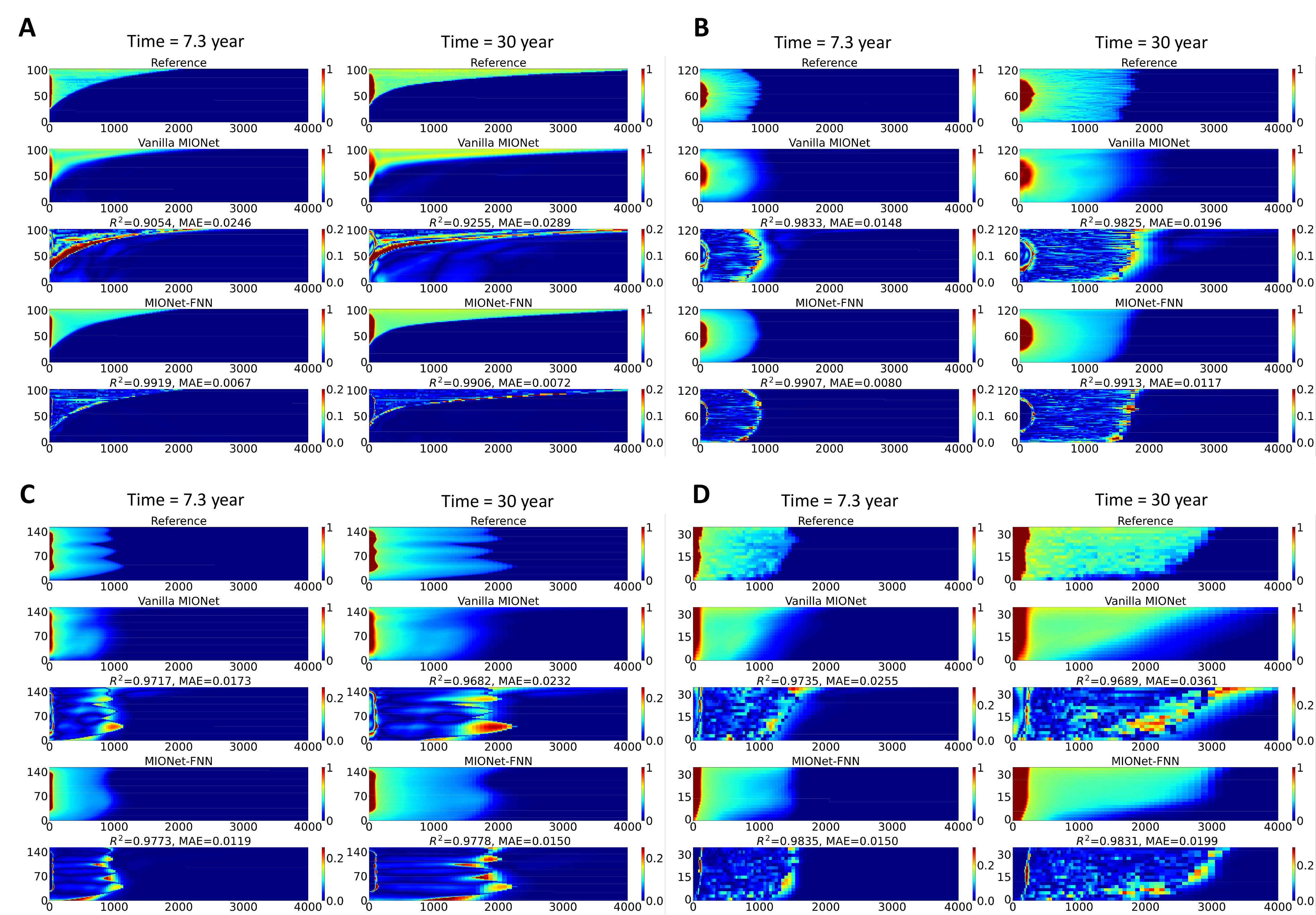}
    \caption{{\textbf{Four testing examples of vanilla MIONet and MIONet-FNN for gas saturation.} For each example, we show the reference solution and the predictions and errors of vanilla MIONet and MIONet-FNN at 7.3 and 30 years. The $R^2$ and MAE error are also computed for each prediction.} }
    \label{fig:mionet_sg}
\end{figure}

We also show 4 examples of the gas saturation prediction and absolute error with the corresponding $R^2$ and MAE in Fig.~\ref{fig:mionet_sg}. Vanilla MIONet and MIONet-FNN have good predictions in the case with smooth solutions (Fig.~\ref{fig:mionet_sg}A). However, when dealing with cases with non-smooth solutions, these two methods result in blurred outputs of interface structures (Figs.~\ref{fig:mionet_sg}B, C, and D). In contrast, Fourier-MIONet generates accurate predictions no matter whether the solution is smooth or not (Fig.~\ref{4 sg cases}).

\section{Conclusions}\label{conclusions}

In this paper, we propose a novel deep neural operator, a Fourier-enhanced multiple-input neural operator (Fourier-MIONet), which combines the advantages of data and computational efficiency from MIONet and good accuracy from U-FNO. We show that Fourier-MIONet is both accurate and efficient for solving the problem of multiphase flow in porous media while ensuring reliability, for safety-critical applications including geological carbon sequestration. Compared with U-FNO, Fourier-MIONet saves 85$\%$ CPU memory and 64$\%$ GPU memory, and can be trained 3.5 times faster. 

Fourier-MIONet also provides much more accurate predictions at unseen time snapshots with only six time snapshots in each case for training. This good generalizability is due to that Fourier-MIONet obeys the physics that the PDE solution is continuous over time, which provides an interpretable basis for maintaining reliability when handling diverse and complex OOD data.  We also consider the nonuniform sampling method to further improve the generalizability.

Here, we give a brief discussion of the time-dependent 3D problem (i.e., 4D problem), since the heterogeneity in a geologic formation is beyond the radially symmetric models. As we do not have a 4D dataset available, we generate a synthetic dataset with the mesh size (100, 100, 18, 24), which means the $x$-axis resolution is 100, $y$-axis resolution is 100, $z$-axis resolution is 18, and there are in total 24 time snapshots. We find that U-FNO is too big to be trained in a GPU with 24 GiB memory, even if the $batch_{case}$ is chosen as 1, and thus we use FNO for the comparison. Fourier-MIONet also uses FNO as the decoder merge net. We have two choices for Fourier-MIONet. First, we only time as the trunk net input and then the merge net should be FNO with 3D FFT. Compared with FNO, Fourier-MIONet (3D FFT) with $batch_{time}$ as 8 only has 10\% the number of parameters and needs much less GPU memory ($<$ 30\%) (Table~\ref{table5:4D estimation}). Second, we can further use $z$-coordinates combined with time as the trunk net input, and then we only need 2D FFT in the merge net. In this case, the number of parameters and GPU memory can be further decreased. The Fourier-MIONet with 2D FFT has 3.3\% number of trainable parameters and 14.7\% GPU memory. The advantages of our architecture show more potential possibilities for 4D multiphase problems, though more work is required to evaluate the performance with 4D datasets in the future.

\begin{table}[htbp]
    \centering
    \caption{\textbf{Comparison between Fourier-MIONet and FNO on time-dependent 3D problems.} We use $batch_{case}$ as 1 for all the experiments.}
    \label{table5:4D estimation}
    \begin{tabular}{lcccc}
        \toprule
        & $batch_{time}$ & $batch_z$&  No. of parameters & GPU memory (GiB)\\ 
        \midrule FNO (4D FFT) & -- & -- &  46,665,329&23.2  \\
        \midrule FMIONet (3D FFT) & \makecell[c]{24 (Full)\\8} &\makecell[c]{--\\--} & \makecell[c]{4,679,033\\4,679,033}& \makecell[c]{16.3\\6.7}\\
        \midrule FMIONet (2D FFT) & 8 & 6 & 1,568,669 & 3.4 \\
        \bottomrule
    \end{tabular}
\end{table}

{{One limitation of the proposed Fourier-MIONet is that it cannot be directly applied to irregular geometry because Fourier and U-Fourier layers require the input to be a regular geometry. There are two possible solutions to this issue. One solution is mapping the irregular geometry to a regular geometry such as gFNO+~\cite{Lu2022deeponetvsfno}. Another solution is using other networks as a merger net, such as MIONet-FNN.}} Another limitation of U-FNO and the proposed Fourier-MIONet is that they cannot provide accurate predictions of the CO\textsubscript{2} saturation plume front (Fig.~\ref{4 sg cases}). One possible solution is adding more Fourier and U-Fourier layers in the decoder merge net, but it will also cost more computing resources. How to improve the prediction of CO\textsubscript{2} plume front will be investigated in future studies.

\section*{Acknowledgments}

This work was supported by ExxonMobil Technology and Engineering Company and U.S. Department of Energy [DE-SC0022953]. The authors gratefully acknowledge Yanhua O. Yuan, Dongzhuo Li, Qiuzi Li, and Alex Gk Lee from ExxonMobil for helpful suggestions. We thank Gege Wen et al. for making the dataset public.

\appendix
{
\section{Network architectures}
{The network architectures of the vanilla MIONet and MIONet-FNN in Section~\ref{subsec:vanilla_mionet} are shown in Tables~\ref{mionet_structure} and \ref{mionet-FNN_structure}, respectively, while the ``CNN'' denotes a convolutional neural network in Table~\ref{CNN_structure}.}}

\begin{table}[htbp]
    \centering
    \caption{{\textbf{Vanilla MIONet architecture}.}}
    \label{mionet_structure}
    \small
    {
    \begin{tabular}{llll}
        \toprule &  & Operation & Output shape\\
        \midrule Branch net & \makecell[l]{Branch net 1\\Branch net 2} & \makecell[l]{CNN\\FNN} & \makecell[l]{($C$, 512)\\($C$, 512)}\\
        \midrule Branch merger operation &  & Point-wise multiplication & ($C$, 512)\\
        \midrule Trunk net &  & FNN & ($T\times96\times200$, 512)\\
        \midrule Branch-Trunk merger operation &  & \makecell[l]{Dot product \\ Reshape}& \makecell[l]{($C$, $T \times 96 \times 200$)\\($C$, $T$, 96, 200)}\\
        \bottomrule
    \end{tabular}}
\end{table}

\begin{table}[htbp]
    \centering
    \caption{{\textbf{MIONet-FNN architecture}.}}
    \label{mionet-FNN_structure}
    \small
    {
    \begin{tabular}{llll}
        \toprule &  & Operation & Output shape\\
        \midrule Branch net & \makecell[l]{Branch net 1\\Branch net 2} & \makecell[l]{CNN\\FNN} & \makecell[l]{($C$, 512)\\($C$, 512)}\\
        \midrule Branch merger operation &  & Point-wise multiplication & ($C$, 512)\\
        \midrule Trunk net &  & FNN & ($T\times96\times200$, 512)\\
        \midrule Branch-Trunk merger operation &  & Point-wise multiplication & ($C\times T \times 96 \times 200$, 512)\\
        \midrule Merger net & & FNN/Reshape & ($C$, $T$, 96, 200)\\
        \bottomrule
    \end{tabular}}
\end{table}

\begin{table}[htbp]
    \centering
    \caption{{\textbf{CNN architecture}. The ``Conv3d'' denotes 3D convolution. ``LeakyReLU'' denotes a leaky rectified linear unit with the negative slope as 0.2.}}
    \label{CNN_structure}
    \small
    {
    \begin{tabular}{clll}
        \toprule Layer & Operation & Output shape\\
        \midrule  1 & Conv3D/LeakyReLU & ($C$, 32, 96, 100)\\
          2 & Conv3D/LeakyReLU & ($C$, 64, 48, 50)\\
          3 & Conv3D/LeakyReLU & ($C$, 64, 48, 50)\\
          4 & Conv3D/LeakyReLU & ($C$, 64, 24, 25)\\
          5 & Conv3D/LeakyReLU & ($C$, 64, 24, 25)\\
          6 & Conv3D/LeakyReLU & ($C$, 128, 12, 13)\\
          7 & Conv3D/LeakyReLU & ($C$, 128, 12, 13)\\
          8 & Conv3D/LeakyReLU & ($C$, 256, 6, 7)\\
          9 & Conv3D/LeakyReLU & ($C$, 256, 6, 7)\\
          10 & Conv3D/LeakyReLU/Flatten & ($C$, 512)\\
          11 & Linear/LeakyReLU & ($C$, 2048)\\
          12 & Linear & ($C$, 512)\\
        \bottomrule
    \end{tabular}}
\end{table}

\bibliographystyle{unsrt}
\bibliography{main}

\end{document}